\documentclass[acmlarge]{acmart}
\usepackage{float}
\usepackage{amsmath,amssymb,amsfonts}
\usepackage{algorithmic}
\usepackage{graphicx}
\usepackage{xcolor}
\usepackage{times}
\usepackage{latexsym}
\usepackage{amssymb}
\usepackage{url}

\newcommand*{\origrightarrow}{}

\renewcommand*{\textrightarrow}{\fontfamily{T1}\selectfont\origrightarrow}

\AtBeginDocument{%
  \providecommand\BibTeX{{%
    \normalfont B\kern-0.5em{\scshape i\kern-0.25em b}\kern-0.8em\TeX}}}

\copyrightyear{2019}
\acmYear{2019}
\setcopyright{acmcopyright}
\acmConference[ACAI '19]{2019 2nd International Conference on Algorithms, Computing and Artificial Intelligence}{December 20--22, 2019}{Sanya, China}
\acmBooktitle{2019 2nd International Conference on Algorithms, Computing and Artificial Intelligence (ACAI '19), December 20--22, 2019, Sanya, China; This is preprint of published manuscript in ACM conference}
\acmPrice{15.00; This is pre-print of Final version published by ACM conference}
\acmDOI{10.1145/3377713.3377774; This is link for Final version}
\acmISBN{978-1-4503-7261-9/19/12}

\begin{document}

\title{An Augmented Translation Technique for low Resource Language pair: Sanskrit to Hindi translation}

\author{Rashi Kumar}
\affiliation{%
	\department{Electronics and Communication Engineering}
	\institution{Malaviya National Institute of Technology}
	\city{Jaipur}
	\country{India}
}
\email{rashi84in@gmail.com}

\author{Piyush Jha}

\affiliation{%
	\department{Electronics and Communication Engineering}
	\institution{Malaviya National Institute of Technology}
	\city{Jaipur}
	\country{India}}
\email{piyushnit15@gmail.com}

\author{Vineet Sahula}
\affiliation{%
		\department{Electronics and Communication Engineering}
	\institution{Malaviya National Institute of Technology}
	\city{Jaipur}
	\country{India}
}
\email{sahula@acm.org}

\begin{abstract}
	Neural Machine Translation (NMT) is an ongoing technique for Machine Translation (MT) using enormous artificial neural network. It has exhibited promising outcomes and has shown incredible potential in solving challenging machine translation exercises. One such exercise is the best approach to furnish great MT to language sets with a little preparing information. In this work, Zero Shot Translation (ZST) is inspected for a low resource language pair. By working on high resource language pairs for which benchmarks are available, namely Spanish to Portuguese, and training on data sets (Spanish-English and English-Portuguese) we prepare a state of proof for ZST system that gives appropriate results on the available data. Subsequently the same architecture is tested for Sanskrit to Hindi translation for which data is sparse, by training the model on English-Hindi and Sanskrit-English language pairs. In order to prepare and decipher with ZST system, we broaden the preparation and interpretation pipelines of NMT seq2seq model in tensorflow, incorporating ZST features. Dimensionality reduction of word embedding is performed to reduce the memory usage for data storage and to achieve a faster training and translation cycles. In this work existing helpful technology has been utilized in an imaginative manner to execute our Natural Language Processing (NLP) issue of Sanskrit to Hindi translation. A Sanskrit-Hindi parallel corpus of 300 is constructed for testing. The data required for the construction of parallel corpus has been taken from the telecasted news, published on Department of Public Information, state government of Madhya Pradesh, India website.                        
	                        
\end{abstract}


\maketitle
\section{Introduction}
Machine Translation (MT) is a fundamental instrument for the interpretation business. Rule-Based MT (RBMT) is the oldest and the most established methodology. The three important steps for the RBMT are text analysis, dictionary mapping, and text generation. Text analysis changes over surface structures into the arrangement of conceivable lexical structures, while text generation changes over a lexical structure into the relating surface structure. Data driven approaches require parallel sentences in source and target languages. The time of Neural Machine Translation have been testing for its improvement, for example, implanting semantic highlights. Most utilized MT ideal models are Phrase-based Statistical Machine Translation (PBSMT)\cite{koehn2007moses} and Neural MT\cite{bahdanau2014neural}. While PBSMT has shown exemplary performance in the scholarly community and also in industry over the last decade, recent studies show that NMT with its incredible potentials has outperformed PBSMT in numerous cases\cite{shterionov2017empirical}.                           
NMT system also rely on large parallel data sets like PBSMT. The interpretation quality relies upon the amount and nature of the preparation corpus accessible. This is a big challenge for language pairs that have sparse parallel data available like Sanskrit-Hindi.                      
In this paper, we overcome the challenge of information sparsity by utilizing zero shot interpretation \cite{Schuster2016zero} for Sanskrit Hindi Language pair. We have developed ZST framework on available parallel data to and 
from English for both the Indian languages. To harness the achievability and capacity of ZST we first structure a proof-of-thought for ZST framework with high-resource tongues (English, Spanish and Portuguese). A ZST framework is then constructed for Sanskrit and Hindi utilizing parallel corpora of Sanskrit-English and English-Hindi to demonstrate the relevance of ZST for scanty information language sets.                         
The quality of translation has been analysed through evaluation of Bilingual Evaluation Understudy (BLEU) score. A test data set is prepared to validate the ZST system that consists of parallel sentences of investigated source and target languages, i.e., Sanskrit-Hindi parallel sentences .                          
The primary commitment of this paper are as follows: Firstly, knowledge gathered from investigation of ZST is utilized for handling the issue associated with machine translation when languages with scanty data set are considered. Secondly, the pipeline of a NMT seq2seq tensorflow model is broadened for implementing ZST.  
The Pretrained fasttext word embedding are used and in order to address the issue of large storage space usage and longer training and testing cycles Principal Component Analysis (PCA) is applied on 300 dimension data set to reduce the dimensions to just half.

The rest of the paper is composed as follows. Section \ref{Sanskrit Hindi Machine Translation Systems} discusses the previous work on Sanskrit Hindi machine translation. Section \ref{Background} presents pertinent foundation and persuades the work done. Section \ref{Zero Shot translation Architecture} explains the architecture designed for implementation of ZST. Section \ref{Result and Discussions} presents results and discussions and Section \ref{Conclusion and Future Direction} concludes the work giving its future perspectives.
\section{Sanskrit Hindi Machine Translation Systems}
\label{Sanskrit Hindi Machine Translation Systems}
The comparisons of the existing and proposed system are shown in Table. \ref{tab:comp}
\begin{itemize}
	\subsection{Sanskrit-Hindi Anusaarka-2009}	
	\item Approach:It is a \textbf{rule based} MT system \cite{bharati2009anusaaraka}.
	\item The insights for the system are taken from Panini's Ashtadhyayi.                 
	\item Developed By:Chinmaya International Foundation (CIF), Indian Institute of Information Technology, Hyderabad (IIIT-H) and University of Hyderabad at Department of Sanskrit Studies.  \item Tool:\textbf{Samsadhani} It is a Language accessor cum machine translation system. Input can be of an of the following encodings Unicode-Devanagari, WX-alphabetic,  Itrans 5.3, Velthuis (VH), Harward Kyoto (KH), Sanskrit Library Project (SLP). Output can be displayed in either Devanagari script or in Roman Diacritical Notation. The system has a Sanskrit language analyser which does the analysis of Sanskrit text using various modules like tokenizer, sandhi splitter, morph analyser, parser, word sense disambiguation, part of speech tagger, chunker, Hindi lexical transfer and a Hindi language generator. End user can see the output of every step of translation.     
	\item The system fails when extended to large domains. It is developed for domains like kids stories, building interactive media and e-learning substance for kids. The proposed system is based on neural machine translation technique that covers all domains in general based on the dataset used.  
	\subsection{Error Analysis of SaHiT}
	\item Approach: It is a \textbf{Statistical Machine Translation}\cite{pandey2016error} and gives a detailed analyses of errors of the system.
	\item Developed by: Scholars at Jawaharlal Nehru University, Delhi at Special Centre for Sanskrit Studies.
	\item Tool:The System is trained on the the Microsoft Translator Hub (MTHub)platform. It can translate simple Sanskrit prose texts. It has a data set of 24k parallel sentences and 25k monolingual sentences. The system gives BLEU scores of 41 and above.       
	\item Sentences that are having compound and sandhi words, for them output accuracy is very low.
    \item Since the system is statistical based it assigns random numbers to two related words, the proposed system uses word embedding that assign close numbers to related words.
\item Statistical Machine Translation (SMT) would evaluate fluency of a sentence in a target language a few words at a time using N-gram language model, the proposed model considers the entire sentence.

Proposed system learn complex relationship between languages as one single model, SMT system have three separate main components-The translation model, reordering model, and the language model.
\end{itemize}
\begin{table*}[htbp]
	\begin{center}
		\small\addtolength{\tabcolsep}{2pt}
		\caption{Comparison of Sanskrit Hindi Translation systems}
		\label{tab:comp}
		\begin{tabular}{|p{5.5cm}|p{5.5cm}|p{5.5cm}|} 
			\hline
			\textbf{Sanskrit-Hindi (Rule Based)} & \textbf{Sanskrit-Hindi (Statistical)} & \textbf{Sanskrit Hindi (Proposed)}\\
			\hline
			The system fails when extended to large domains. It is developed for domains like kids stories, building interactive media and e-learning substance for kids. & Sanskrit-Hindi text corpora has been collected or developed manually from the literature, health, news and tourism domains. & The proposed system is based on neural machine translation technique that covers all domains in general based on the dataset used.\\
			\hline
			Hand crafted rules based on Panini's Ashtadhyayi. & Based on Bayes theorem. Similar words assigned random numbers. & Word embedding used, similar words have close numbers.\\
			\hline
			Separate modules like tokenizer, sandhi splitter, morph analyser, parser, word sense disambiguation, part of speech tagger, chunker, Hindi lexical transfer and a Hindi language generator used to get translation. & 
	        SMT system have three separate main components-The translation model, reordering model, and the language model.SMT would evaluate fluency of a sentence in a target language a few words at a time using N-gram language model.	& Proposed system learns complex relationship between languages as one single model. Proposed model considers the entire sentence.\\
	 \hline
	 \end{tabular}
	\end{center}
\end{table*}
\section{Proposed Technique for Sanskrit Hindi Machine Translation}
\label{Background}
This section gives the details of the ZST approach and various modules and algorithms used in the implementation of the work done. 
\subsection{Zero-shot translation}
Zero-shot Translation (ZST) \cite{Schuster2016zero} is a way to deal with training of a solitary NMT system to decipher between various dialects. In such a system the pair of source and target language sentences are not seen by the system during training but still it provides interpretation to an objective language given a source language as an input. ZST method avoids rebuilding of a system for every new language pair that it sees. Authors in \cite{Schuster2016zero} propose that a NMT system can be extended to a multiple dialect ZST system by appending a token before every source sentence expressing the objective dialect. In particular, a token $<$2T1$>$ will be appended towards the start of the source sentence which is say language S1 and has to be changed to a target sentence in language T1. In the \cite{Schuster2016zero}, \cite{ha2016toward} a solitary shared attention module with a solitary 'general' encoder-decoder for overall dialects is utilized. \cite{firat2016zero} Likewise present a multilingual methodology that utilize a solitary consideration module however numerous encoders and decoders for different source and target dialects. We have focused on the single encoder and decoder model with shared attention module. In order to implement the approach, progressions are made to the interpretation pipelines of NMT seq2seq model in Tensorflow and some of the pre-processing steps. We augment the works of \cite{firat2016zero} and \cite{Schuster2016zero}. We apply zero shot interpretation for one specific language pair Sanskrit- Hindi with a solitary encoder and decoder module and a common attention instrument, while utilizing Hindi-English and English-Sanskrit language sets.                
\subsection{Word Embedding}
Word Embedding step is able to catch setting of a word in a report concerning semantic and syntactic similitude in connection with different words. 
The words with comparable setting will possess close spatial positions. If the cosine of angle between similar vectors is found to be near to 1, it means the angle is very near to 0. Lets take an example on Cosine Similarity\\

Let A="apple, banana, banana"=$<$1,2,0$>$, B="apple, banana, grapes, grapes"=$<$1,1,2$>$ \newline
query has 3 dimensions\newline
\begin{equation}\label{cosinesimilarity}
\sigma(X,Y)=\frac{1*1+2*1+2*0}{\sqrt(1^2+2^2+0^2)*\sqrt(1^2+1^2+2^2)}= 0.55
\end{equation}
if we compare A with itself then $\sigma$(A,A)=1.
Word2Vec approach is one such strategy to build such an inserting. It may be applied utilizing two strategies, which use neural networks. Skip Gram and Common Bag Of Words (CBOW). Word2vec has been used in $ZST_{1}$. In $ZST_{2}$ fasttext pretrained word vectors are used. FastText library is developed by Facebook's AI Research and can be used for word embedding learning and classification of text.\newline
Size of embedding matrix($Size_{emb.mat}$) is obtained as the product of number of words($N_{w}$) and number of hidden units($N_{hu}$)
\begin{equation}\label{size of embedding matrix}
Size_{emb.mat} = N_{w} * N_{hu}
\end{equation}
\subsection{Long Short Term Memory Encoder Decoder}
\begin{figure}[H]
	\centering
	\includegraphics[width=8.5cm,height=4.5cm]{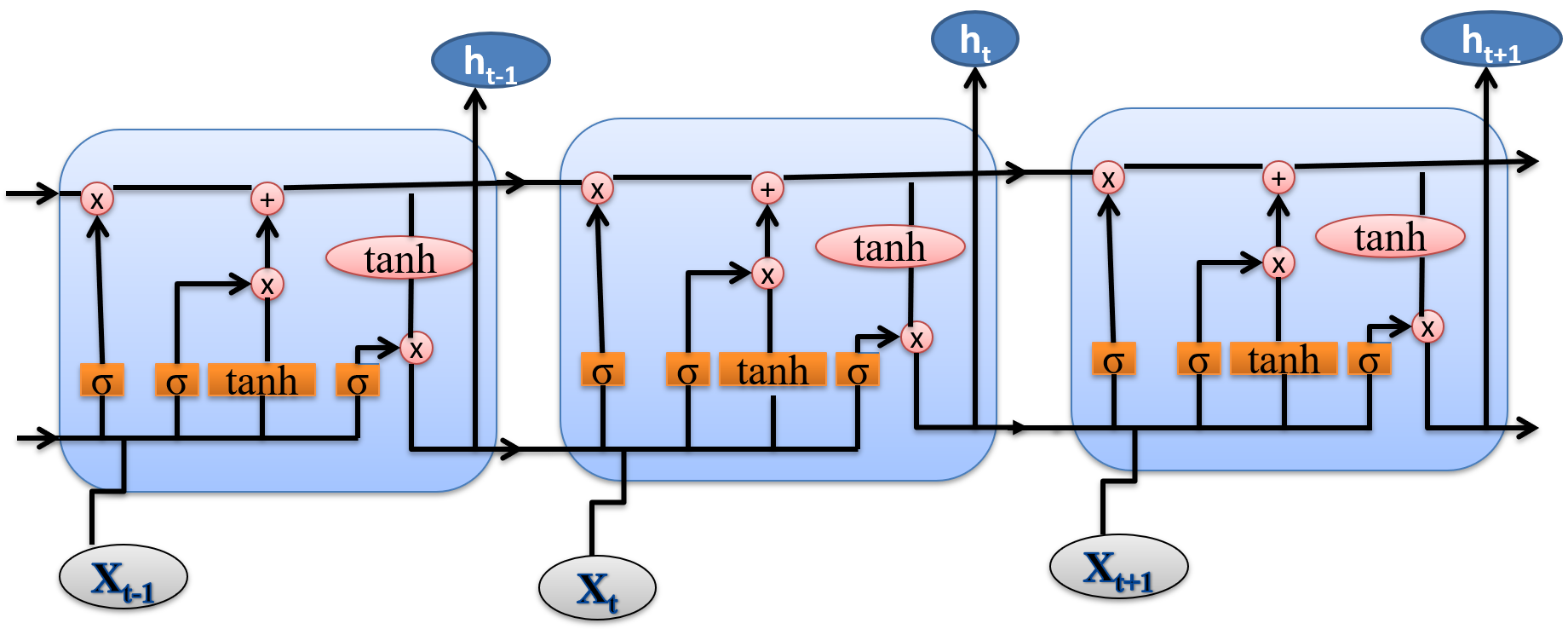}
	\caption{LSTM Structure}
	\label{ls}
\end{figure}

Long Short Term Memory Encoder Decoder (LSTM) \cite{hochreiter1997long} are a sort of Recurrent Neural Network (RNN) which are skilled to recollect data for extensive stretches of time. LSTM's also have this chain-like structure like the RNN's, yet the structure of the rehashing module is not quite the same as RNN.   Rather than a single neural framework layer, there are four layers in a module. The \textit{sigmoid} and \textit{tanh} layers interface inside a similar module also with various modules for learning are used Fig.\ref{ls}. shows the LSTM structure.
\vspace{-0.5ex}
\vspace{-0.5ex}
In our work we use four bidirectional LSTM layers stacked one above the other.  This architecture depends on the idea that the yield at any time moment may rely upon past information as well as on future information. Utilizing this thought, the LSTM is changed to associate two concealed layers of inverse bearings to a similar yield. This changed variant of LSTM is known as a Bidirectional LSTM(Bi-LSTM)\cite{schuster1997bidirectional}.               
\subsection{Attention module}
\begin{figure}[h]
	\centering
	\includegraphics[width=8cm,height=6cm]{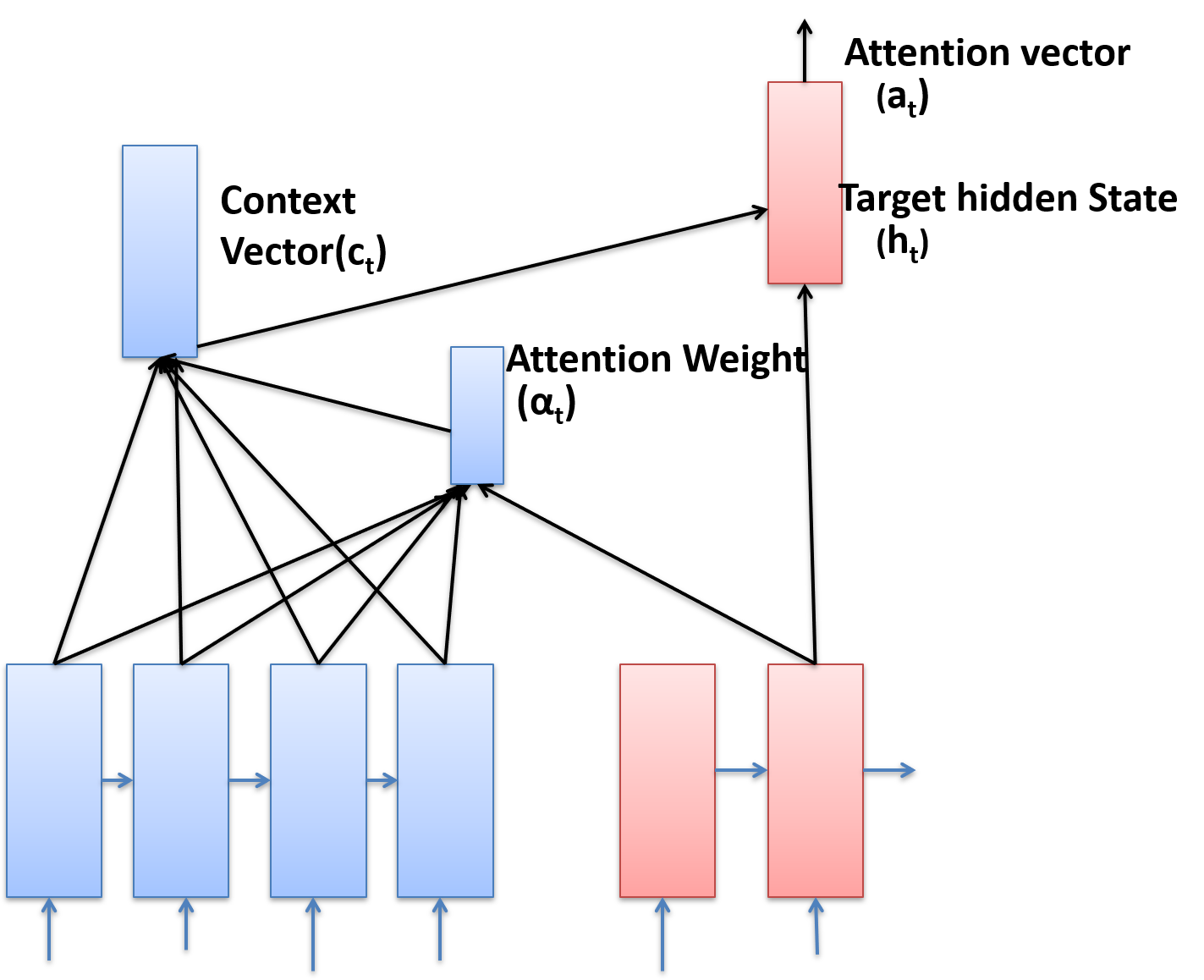}
	\caption{Attention Module \cite{luong17}}
	\label{attention}
\end{figure}

The attention calculation occurs at each decoder time step \cite{luong17}. The present target hidden state is contrasted with all source states to determine the attention weight. The scoring function (Score()) can have the multiplicative or additive forms (Luong's multiplicative style and Bahdanau's additive  style)\cite{luong2015effective}. We have used the additive form. Score function is used to compare the target hidden state  with every hidden state of the source. The normalized result produce attention weights. The module yielding the attention vector along with its intermediate stages is shown in Fig.\ref{attention}.
\vspace{-0.5ex}
\vspace{-0.5ex}
\subsection{Adaptive Moment Estimation optimization Algorithm}
Adaptive Moment Estimation (Adam) optimization algorithm with default learning rate of 0.001, beta$_1$=0.9, beta$_2$=0.999, $epsilon=1*\exp(-7)$, amsgrad=False, is used in place of the stochastic gradient descent technique, and system weights are changed iteratively depending on training data, as well as on the estimation of first and second order moments\cite{kingma2014adam}. Learning rate is a Tensor or a floating point value. Beta$_1$ is a float value or a constant float tensor. The exponential decay rate for the 1st moment estimates. Beta$_2$ is a float value or a constant float tensor. The exponential decay rate for the 2nd moment estimates. Epsilon is a small constant for numerical stability. Amsgrad is a Boolean \cite{reddi2019convergence}. Algorithm has the benefits of both Adaptive Gradient Algorithm(AdaGrad) and Root Mean Square Propagation(RMSProp).                     
\subsection{Loss Function}
\begin{figure}[H]
	\centering
	\includegraphics[width=6cm,height=6cm]{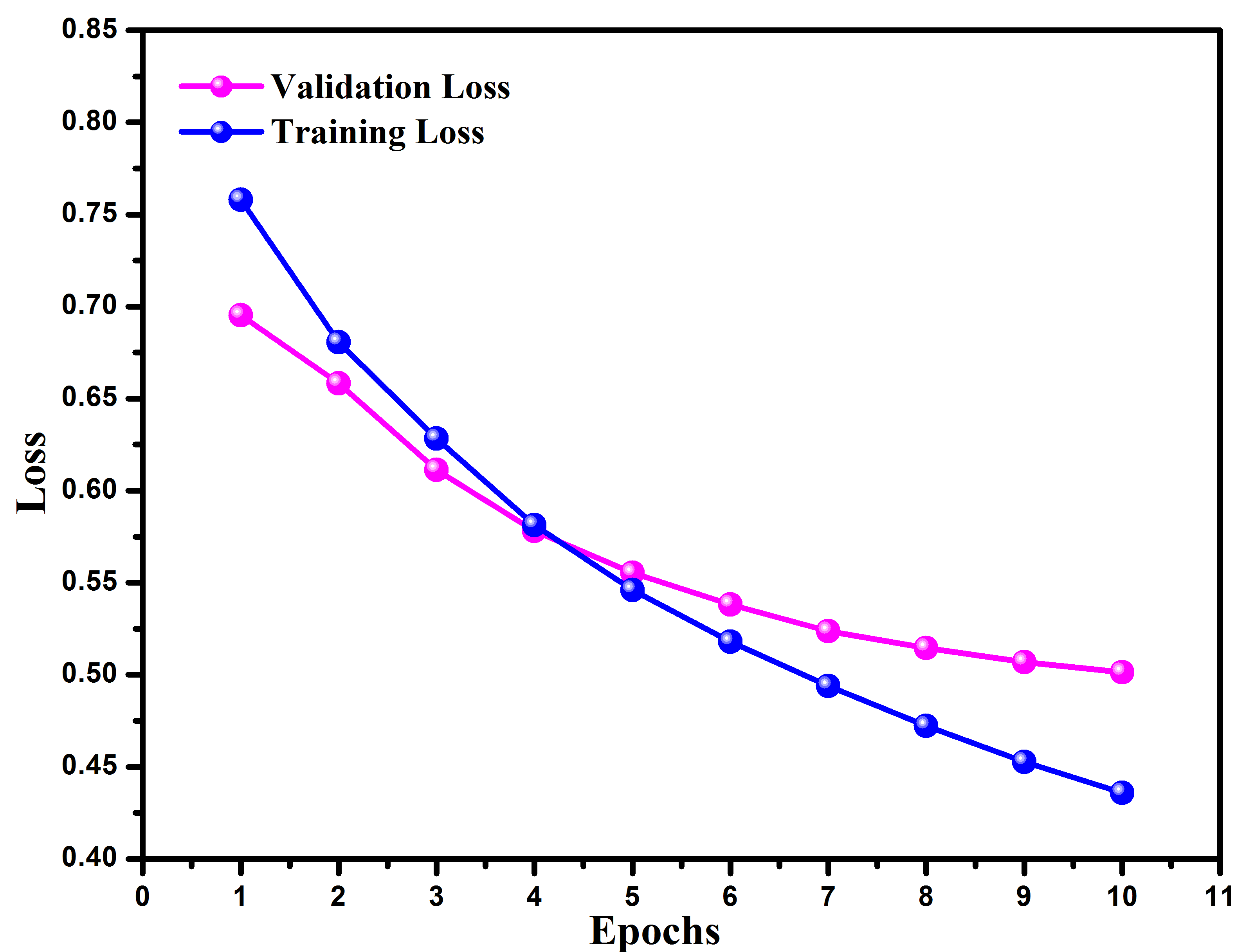}
	\caption{Loss vs Epochs}
	\label{loss}
\end{figure}

Loss function is calculated using tensorflow API.
\begin{equation}
Cross EntropyLoss=-\sum_{w=1}^{|S|}\sum_{e=1}^{|V|}y_{w,e}\log({\hat{y}_{w,e})} 
\end{equation}
$|S|$=Sentence Length\newline
$|V|$=Vocabulary Length\newline
$\hat{y}_{w,e}$= Given word w, probability predicted of the vocab entry e.\newline
$y_{w,e}$=1, when entry of vocabulary is correct word.\newline
$y_{w,e}$=0, when entry of vocabulary is not correct word.\newline
This provides the softmax cross entropy computation between actual and predicted values. Cross entropy loss indicates the separation between actual and predicted outputs. Fig. \ref{loss} illustrates the Loss versus Epochs plot. 
\vspace{-0.5ex}

Softmax is a kind of activation layer that interprets the outputs as probabilities usually used in the last layer of the neural network.
\subsection{Training}
The input is passed through the encoder and it returns encoder output and the encoder hidden state. The decoder input which is the start token, encoder output and its hidden state are passed to the decoder. The predicted output from the decoder and its hidden state are available at the decoder end. The predicted output is used to calculate the loss and the hidden state is fed back to the decoder. Teacher forcing, a technique where the target word is sent as the next input, is used to feed the next input to the decoder. In the last step gradient is calculated and applied to the optimizer and back propagate thereafter.     \subsection{Translate}
The translation is similar to training except teacher forcing is not used instead predicted word from previous stage together with hidden state and encoder output are fed to decoder at each time instance. When the decoder receives the end token, prediction stops there and attention weights are stored at each time instance.                          
\subsection{Decoding Method -Greedy Search Decoding}
We utilize greedy search that chooses the most appropriate word at each progression in the yield  sequence.                          
The argmax() numerical function is utilized to choose the index of an array that has the biggest value. This function is given in numpy library. The sequence of integers that represent the index of the words that have the largest value can then be mapped back to words in the corpus.
\section{Zero Shot translation Architecture}
\label{Zero Shot translation Architecture}
The architecture of ZST multilingual translation system is similar to Googles NMT system \cite{johnson2017google}. We have adapted and augmented the architecture by adding a token at the start of sentence indicating the target language. A NMT architecture utilizes a multilayered LSTM to outline input information succession to a vector of a fixed dimensionality, and after that uses another profound LSTM to unravel the objective grouping from the vector\cite{sutskever2014sequence}.                          
A parallel corpus of Sanskrit-Hindi is constructed for testing. The data required for the construction of parallel corpus has been taken from the news published on Department of Public Information Madhya Pradesh (M.P),India government website. As per our knowledge this is the first implementation of Sanskrit-Hindi translation by adding ZST capabilities to the sequence2sequence NMT model.      
\subsection {System Engines}
There are two ZST multilingual system engines produced in this work. One is for Spanish Portuguese translation using Spanish-English and English-Portuguese data sets. This engine acts as a proof of concept for developing the Sanskrit-Hindi translation system which uses Parallel corpus's of Sanskrit-English and English-Hindi language pairs.           \vspace{-1ex}               
\begin{table}[H]
	\begin{center}
		\small\addtolength{\tabcolsep}{2pt}
		\caption{System Engines}
		\label{tab:table1}
		\begin{tabular}{|p{1cm}|p{1.8cm}|c|} 
			\hline
			\textbf{Engine Name} & \textbf{Language Used} & \textbf{Used for Translating }\\
			\hline
			$ZST_{1}$ & SP-EN, EN-PR  & SP-PR\\
			\hline
			$ZST_{2}$ & SA-EN, EN-HI  & SA-HI\\
			\hline
		\end{tabular}
	\end{center}
\end{table}
\vspace{-1ex}

For the first system engine we have used pre-trained word vectors from word2vec model \cite{mikolov2013distributed}.                          
In the second system engine we use Fasttext pre-trained word vector representation for Sanskrit, English and Hindi languages. Fasttext models are prepared utilizing continuous bag of words(CBOW) with position-loads of 300 dimension, character n-gram having length of 5, size of window 5 and with negatives to be 10\cite{grave2018learning}. Keras Libraries are used to make the encoder and decoder of the translational model.
\vspace{-0.5ex}
\begin{figure}[H]
	\centering
	\includegraphics[width=8cm,height=4cm]{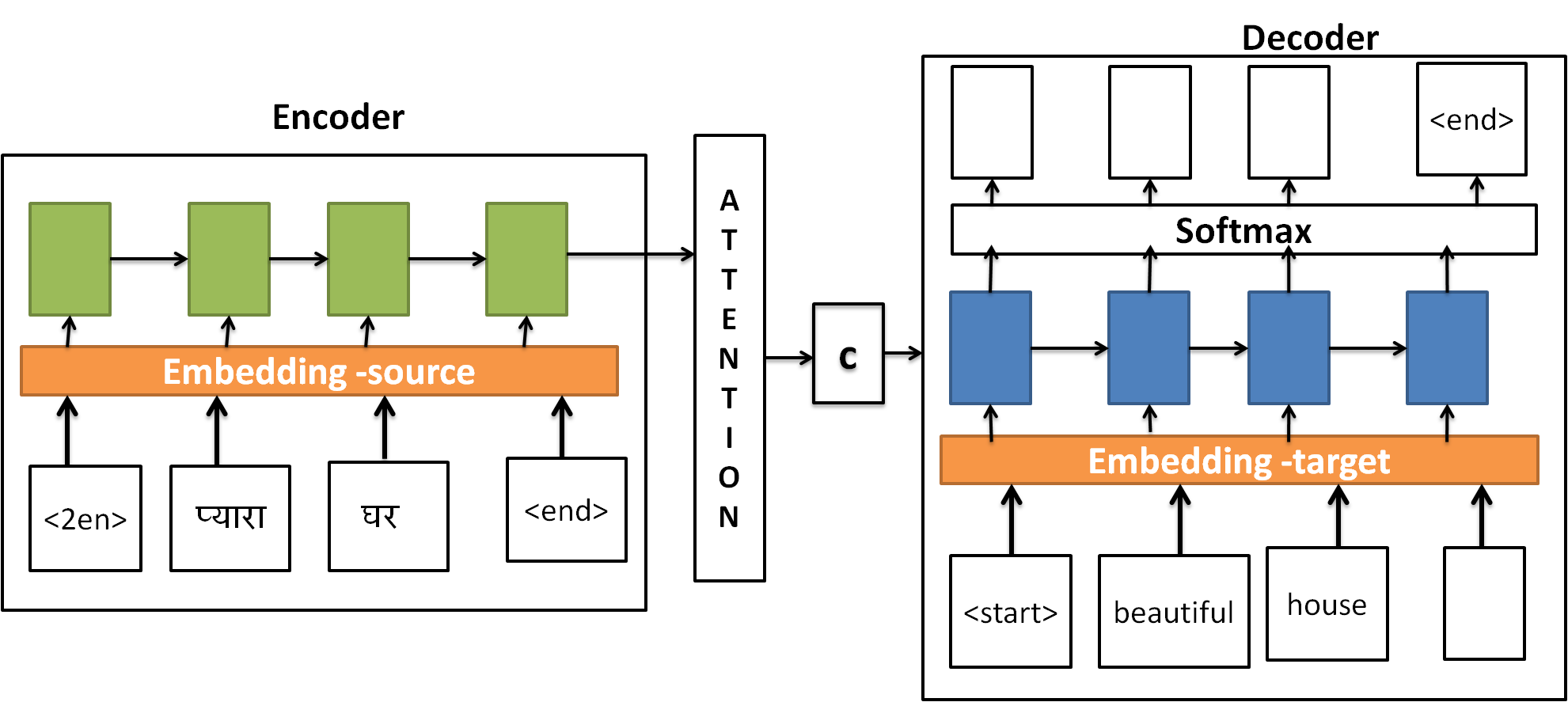}
	\caption{Basic ZST Block}
	\label{ZSTblock}
\end{figure}
\vspace{-0.5ex}
The first step when working with text data is preprocessing of the data. As the information accessible is unstructured and noisy, to accomplish better outcomes and to fabricate better calculations it is important to clean the information. The basic module of ZST is shown in Fig.\ref{ZSTblock}

\begin{table}[]
	\begin{center}
		\small\addtolength{\tabcolsep}{2pt}
		\caption{Data sets Used}
		\label{tab:table2}
		\begin{tabular}{|p{3cm}|p{1.5cm}|p{3cm}|} 
			\hline
			\textbf{Language pairs} & \textbf{No. of sentences} & \textbf{Source of data set}\\
			\hline
			Portuguese-English & 148713 & Tatoeba \cite{EN-Po}\\
			\hline
			Spanish-English & 122936 & Tatoeba \cite{EN-Sp}\\
			\hline
			Portuguese-spanish & 392 & https://data.europa.eu \\
			\hline
			English-Hindi & 208850 & IIT Bombay, India \cite{kunchukuttan2017iit}\\
			\hline
			Sanskrit-English & 13463 & IIT Kanpur, India \cite{sa-en1}, \cite{sa-en2}\\
			\hline
		\end{tabular}
	\end{center}
\end{table}
\vspace{-0.5ex}

\begin{table}[H]
	\begin{center}
		\small\addtolength{\tabcolsep}{2pt}
		\caption{Self Generated Corpuses for Testing}
		\label{tab:table3}
		\def\arraystretch{1.5}
		\begin{tabular}{|p{1.2cm}|p{6.3cm}|} 
			\hline
			\textbf{Language pair} & \textbf{source of data set}\\
			\hline
			Hindi-Sanskrit  & Test data set of 300 sentences prepared from (i) news published on Department of Public Information, Bhopal website (ii) "Manogatam" Sanskrit translation website of All India Radio Show (iii) "Mann Ki Baat" speeches.\\
			\hline
		\end{tabular}
	\end{center}
\end{table}
\vspace{-2ex}
\subsection{Reduction of Embedding Dimensions}
\begin{figure*}[htbp]
	\centering
	\includegraphics[width=0.9\textwidth]{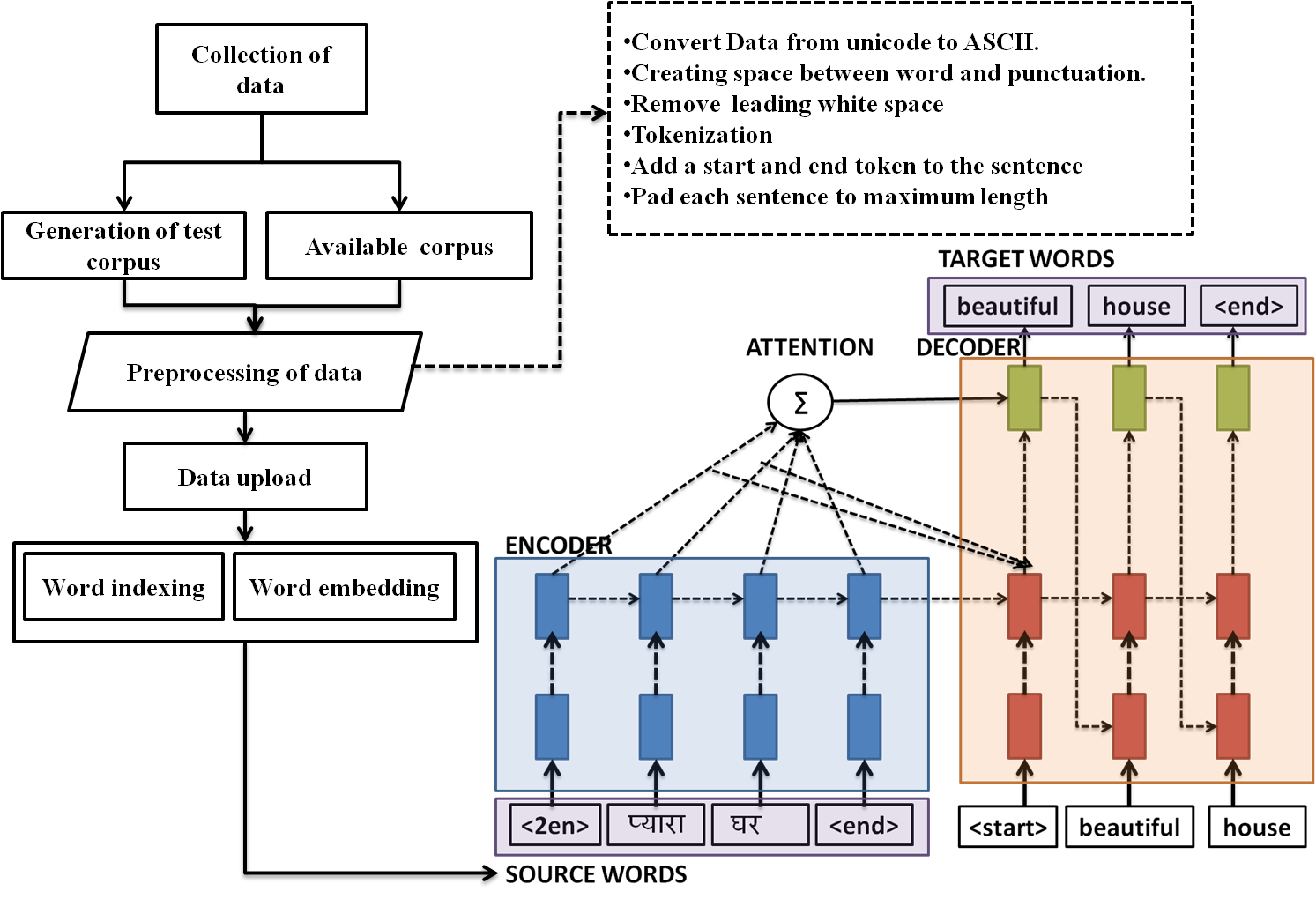}
	\caption{Process Flow}
	\label{flow}
\end{figure*}

Pre-trained Word data set embedding of 300 dimensions have been used as the basic block for text processing on both the encoder and decoder side. The word embedding size is decreased by reducing the dimensions. This makes the use of word data set embedding feasible in devices that have constrained on their memory. Principal Component Analysis (PCA) based dimensionality decrease with some post processing steps have been utilized to develop word data set embedding of lower dimensions of 150. Translation results are produced in Table. \ref{tab:Bleu}.
\vspace{-0.5ex}
\begin{table}[H]
	\begin{center}
		\small\addtolength{\tabcolsep}{2pt}
		\caption{BLEU Score comparison}
		\centering
		\label{tab:Bleu}
		\def\arraystretch{1.5}
		\begin{tabular}{|p{6cm}|c|} 
			\hline
			\textbf{Zero-shot (1-directional)} & \textbf{BLEU Score} \\
			\hline
			Model-1:(no correlation loss, word2vec, no bidirectional LSTM) & 8.2 \\
			\hline
			Model-2:(correlation loss, word2vec, no bidir LSTM) & 8.2 \\
			\hline
			Model-3:(correlation loss, word2vec, bidir LSTM) & 12.0 \\
			\hline
			Model-4:(Model-3 + basic hindi stemming) & 13.3\\
			\hline
		\end{tabular}
	\end{center}
\end{table}
\vspace{-1ex}
We have implemented the system by using core algorithms from \cite{luong17} 
The computing environment consists of high performance GPU, Tesla V100. 16GB Memory, 61 GB RAM and 100 GB SSD.
The detailed process flow is shown in Fig.\ref{flow}.        
\vspace{-4ex} 
\section{Result and Discussions}
\label{Result and Discussions}
The system has been evaluated using BLEU scores. Initially the system is evaluated without providing the word embedding on a data set with number of sentences limited to 10,000 and system is trained for 10 Epochs. The scores obtained are low due to the limitation of costly computational resources. While training we limit the number of sentences for training as it takes too much of time for the system to get trained on a big corpus. The number of units are also limited to 128 in LSTM. Later the system is evaluated with pre-trained word embedding which led to improvement of results with the same hyper parameter settings. If we increase number of LSTM units, number of epochs for which system is trained and also provide a big data set, the results would improve. As the number of epochs were increased 20 the BLEU scores improved. Further increasing the number of epochs to 30 deteriorated the scores. Now the number of sentences that are provided while training have to be increased for which work on corpus development has to be done.     

\section{Conclusions and Future Directions}
\label{Conclusion and Future Direction}
The BLEU scores obtained are low as the number of sentences in Sanskrit-English parallel corpus were reduced to 6000 after pre-processing which deleted sentences that had more than thirty words in a single line. If the system was trained with sentences having more than thirty words the system encountered out of memory error. Work on development of a good Sanskrit-English parallel corpus is due. The PCA based dimensionality reduction significantly lessens words data set embedding size while enhancing their use having a straightforward execution. It has allowed the utilization of word data set embedding on devices that have constrained memory usage by fundamentally reducing the memory necessities.

Presently, Adam optimization with default learning rate of 0.001 has been used. This can aid in assessment of the system performance for different algorithms whose BLEU scores can be compared in order to select the best performing candidate for real life applications. We have employed one way engines which entail reduce time and process overheads. However, implementation of two-way engines is necessary on which we are currently working. The performance of the implemented system can further be enhanced by use of extensive pre-processing steps related to cleaning of data, like stemming. Performing linguistics analysis of Sanskrit data by morphological segmentation of compound and conjunction word will lead to further improvement of scores. The morphological segmentation be rule based or the data driven, these modules could be integrated with ZST model.
\vspace{-1.5ex}
\subsection*{Acknowledgment} This work was supported by Visvesvaraya PhD scheme and SMDP-C2SD project of Ministry of Electronics \& IT, Government of India, CPIN (MNITJ)- 1000110042.
\vspace{-1.5ex}
\bibliographystyle{ACM-Reference-Format}
\bibliography{zst}


\begin{thebibliography}{24}


\ifx \showCODEN    \undefined \def \showCODEN     #1{\unskip}     \fi
\ifx \showDOI      \undefined \def \showDOI       #1{#1}\fi
\ifx \showISBNx    \undefined \def \showISBNx     #1{\unskip}     \fi
\ifx \showISBNxiii \undefined \def \showISBNxiii  #1{\unskip}     \fi
\ifx \showISSN     \undefined \def \showISSN      #1{\unskip}     \fi
\ifx \showLCCN     \undefined \def \showLCCN      #1{\unskip}     \fi
\ifx \shownote     \undefined \def \shownote      #1{#1}          \fi
\ifx \showarticletitle \undefined \def \showarticletitle #1{#1}   \fi
\ifx \showURL      \undefined \def \showURL       {\relax}        \fi
\providecommand\bibfield[2]{#2}
\providecommand\bibinfo[2]{#2}
\providecommand\natexlab[1]{#1}
\providecommand\showeprint[2][]{arXiv:#2}

\bibitem[\protect\citeauthoryear{Bahdanau, Cho, and Bengio}{Bahdanau
  et~al\mbox{.}}{2014}]%
        {bahdanau2014neural}
\bibfield{author}{\bibinfo{person}{Dzmitry Bahdanau},
  \bibinfo{person}{Kyunghyun Cho}, {and} \bibinfo{person}{Yoshua Bengio}.}
  \bibinfo{year}{2014}\natexlab{}.
\newblock \showarticletitle{Neural machine translation by jointly learning to
  align and translate}.
\newblock \bibinfo{journal}{\emph{arXiv preprint arXiv:1409.0473}}
  (\bibinfo{year}{2014}).
\newblock


\bibitem[\protect\citeauthoryear{Bharati and Kulkarni}{Bharati and
  Kulkarni}{2009}]%
        {bharati2009anusaaraka}
\bibfield{author}{\bibinfo{person}{Akshar Bharati} {and} \bibinfo{person}{Amba
  Kulkarni}.} \bibinfo{year}{2009}\natexlab{}.
\newblock \showarticletitle{Anusaaraka: An Accessor cum Machine Translator}.
\newblock \bibinfo{journal}{\emph{Department of Sanskrit Studies, University of
  Hyderabad, Hyderabad}} (\bibinfo{year}{2009}).
\newblock


\bibitem[\protect\citeauthoryear{Firat, Sankaran, Al-Onaizan, Vural, and
  Cho}{Firat et~al\mbox{.}}{2016}]%
        {firat2016zero}
\bibfield{author}{\bibinfo{person}{Orhan Firat}, \bibinfo{person}{Baskaran
  Sankaran}, \bibinfo{person}{Yaser Al-Onaizan}, \bibinfo{person}{Fatos
  T~Yarman Vural}, {and} \bibinfo{person}{Kyunghyun Cho}.}
  \bibinfo{year}{2016}\natexlab{}.
\newblock \showarticletitle{Zero-resource translation with multi-lingual neural
  machine translation}.
\newblock \bibinfo{journal}{\emph{arXiv preprint arXiv:1606.04164}}
  (\bibinfo{year}{2016}).
\newblock


\bibitem[\protect\citeauthoryear{Gitasupersite.iitk.ac.in}{Gitasupersite.iitk.ac.in}{2019}]%
        {sa-en1}
\bibfield{author}{\bibinfo{person}{Gitasupersite.iitk.ac.in}.}
  \bibinfo{year}{2019}\natexlab{}.
\newblock \bibinfo{booktitle}{\emph{Gita Supersite}}.
\newblock
\urldef\tempurl%
\url{https://www.gitasupersite.iitk.ac.in/srimad?language=dv&field_chapter_value=1&field_nsutra_value=1}
\showURL{%
Retrieved November 23, 2019 from \tempurl}


\bibitem[\protect\citeauthoryear{Grave, Bojanowski, Gupta, Joulin, and
  Mikolov}{Grave et~al\mbox{.}}{2018}]%
        {grave2018learning}
\bibfield{author}{\bibinfo{person}{Edouard Grave}, \bibinfo{person}{Piotr
  Bojanowski}, \bibinfo{person}{Prakhar Gupta}, \bibinfo{person}{Armand
  Joulin}, {and} \bibinfo{person}{Tomas Mikolov}.}
  \bibinfo{year}{2018}\natexlab{}.
\newblock \showarticletitle{Learning Word Vectors for 157 Languages}. In
  \bibinfo{booktitle}{\emph{Proceedings of the International Conference on
  Language Resources and Evaluation {(LREC 2018)}}}.
\newblock


\bibitem[\protect\citeauthoryear{Ha, Niehues, and Waibel}{Ha
  et~al\mbox{.}}{2016}]%
        {ha2016toward}
\bibfield{author}{\bibinfo{person}{Thanh-Le Ha}, \bibinfo{person}{Jan Niehues},
  {and} \bibinfo{person}{Alexander Waibel}.} \bibinfo{year}{2016}\natexlab{}.
\newblock \showarticletitle{Toward multilingual neural machine translation with
  universal encoder and decoder}.
\newblock \bibinfo{journal}{\emph{arXiv preprint arXiv:1611.04798}}
  (\bibinfo{year}{2016}).
\newblock


\bibitem[\protect\citeauthoryear{Hochreiter and Schmidhuber}{Hochreiter and
  Schmidhuber}{1997}]%
        {hochreiter1997long}
\bibfield{author}{\bibinfo{person}{Sepp Hochreiter} {and}
  \bibinfo{person}{J{\"u}rgen Schmidhuber}.} \bibinfo{year}{1997}\natexlab{}.
\newblock \showarticletitle{Long short-term memory}.
\newblock \bibinfo{journal}{\emph{Neural computation}} \bibinfo{volume}{9},
  \bibinfo{number}{8} (\bibinfo{year}{1997}), \bibinfo{pages}{1735--1780}.
\newblock


\bibitem[\protect\citeauthoryear{Johnson, Schuster, Le, Krikun, Wu, Chen,
  Thorat, Vi{\'e}gas, Wattenberg, Corrado, et~al\mbox{.}}{Johnson
  et~al\mbox{.}}{2017}]%
        {johnson2017google}
\bibfield{author}{\bibinfo{person}{Melvin Johnson}, \bibinfo{person}{Mike
  Schuster}, \bibinfo{person}{Quoc~V Le}, \bibinfo{person}{Maxim Krikun},
  \bibinfo{person}{Yonghui Wu}, \bibinfo{person}{Zhifeng Chen},
  \bibinfo{person}{Nikhil Thorat}, \bibinfo{person}{Fernanda Vi{\'e}gas},
  \bibinfo{person}{Martin Wattenberg}, \bibinfo{person}{Greg Corrado},
  {et~al\mbox{.}}} \bibinfo{year}{2017}\natexlab{}.
\newblock \showarticletitle{Google\textquotesingle s multilingual neural
  machine translation system: Enabling zero-shot translation}.
\newblock \bibinfo{journal}{\emph{Transactions of the Association for
  Computational Linguistics}}  \bibinfo{volume}{5} (\bibinfo{year}{2017}),
  \bibinfo{pages}{339--351}.
\newblock


\bibitem[\protect\citeauthoryear{Kelly}{Kelly}{2019a}]%
        {EN-Po}
\bibfield{author}{\bibinfo{person}{C. Kelly}.}
  \bibinfo{year}{2019}\natexlab{a}.
\newblock \bibinfo{booktitle}{\emph{English-Portuguese Sentences from the
  Tatoeba Project}}.
\newblock
\urldef\tempurl%
\url{https://www.manythings.org/bilingual/por/}
\showURL{%
Retrieved November 23, 2019 from \tempurl}


\bibitem[\protect\citeauthoryear{Kelly}{Kelly}{2019b}]%
        {EN-Sp}
\bibfield{author}{\bibinfo{person}{C. Kelly}.}
  \bibinfo{year}{2019}\natexlab{b}.
\newblock \bibinfo{booktitle}{\emph{English-Spanish Sentences from the Tatoeba
  Project}}.
\newblock
\urldef\tempurl%
\url{https://www.manythings.org/bilingual/spa/}
\showURL{%
Retrieved November 23, 2019 from \tempurl}


\bibitem[\protect\citeauthoryear{Kingma and Ba}{Kingma and Ba}{2014}]%
        {kingma2014adam}
\bibfield{author}{\bibinfo{person}{Diederik~P Kingma} {and}
  \bibinfo{person}{Jimmy Ba}.} \bibinfo{year}{2014}\natexlab{}.
\newblock \showarticletitle{Adam: A method for stochastic optimization}.
\newblock \bibinfo{journal}{\emph{arXiv preprint arXiv:1412.6980}}
  (\bibinfo{year}{2014}).
\newblock


\bibitem[\protect\citeauthoryear{Koehn, Hoang, Birch, Callison-Burch, Federico,
  Bertoldi, Cowan, Shen, Moran, Zens, et~al\mbox{.}}{Koehn
  et~al\mbox{.}}{2007}]%
        {koehn2007moses}
\bibfield{author}{\bibinfo{person}{Philipp Koehn}, \bibinfo{person}{Hieu
  Hoang}, \bibinfo{person}{Alexandra Birch}, \bibinfo{person}{Chris
  Callison-Burch}, \bibinfo{person}{Marcello Federico}, \bibinfo{person}{Nicola
  Bertoldi}, \bibinfo{person}{Brooke Cowan}, \bibinfo{person}{Wade Shen},
  \bibinfo{person}{Christine Moran}, \bibinfo{person}{Richard Zens},
  {et~al\mbox{.}}} \bibinfo{year}{2007}\natexlab{}.
\newblock \showarticletitle{Moses: Open source toolkit for statistical machine
  translation}. In \bibinfo{booktitle}{\emph{Proceedings of the 45th annual
  meeting of the association for computational linguistics companion volume
  proceedings of the demo and poster sessions}}. \bibinfo{pages}{177--180}.
\newblock


\bibitem[\protect\citeauthoryear{Kunchukuttan, Mehta, and
  Bhattacharyya}{Kunchukuttan et~al\mbox{.}}{2017}]%
        {kunchukuttan2017iit}
\bibfield{author}{\bibinfo{person}{Anoop Kunchukuttan}, \bibinfo{person}{Pratik
  Mehta}, {and} \bibinfo{person}{Pushpak Bhattacharyya}.}
  \bibinfo{year}{2017}\natexlab{}.
\newblock \showarticletitle{The iit bombay english-hindi parallel corpus}.
\newblock \bibinfo{journal}{\emph{arXiv preprint arXiv:1710.02855}}
  (\bibinfo{year}{2017}).
\newblock


\bibitem[\protect\citeauthoryear{Luong, Brevdo, and Zhao}{Luong
  et~al\mbox{.}}{2017}]%
        {luong17}
\bibfield{author}{\bibinfo{person}{Minh{-}Thang Luong}, \bibinfo{person}{Eugene
  Brevdo}, {and} \bibinfo{person}{Rui Zhao}.} \bibinfo{year}{2017}\natexlab{}.
\newblock \showarticletitle{Neural Machine Translation (seq2seq) Tutorial}.
\newblock \bibinfo{journal}{\emph{https://github.com/tensorflow/nmt}}
  (\bibinfo{year}{2017}).
\newblock


\bibitem[\protect\citeauthoryear{Luong, Pham, and Manning}{Luong
  et~al\mbox{.}}{2015}]%
        {luong2015effective}
\bibfield{author}{\bibinfo{person}{Minh-Thang Luong}, \bibinfo{person}{Hieu
  Pham}, {and} \bibinfo{person}{Christopher~D Manning}.}
  \bibinfo{year}{2015}\natexlab{}.
\newblock \showarticletitle{Effective approaches to attention-based neural
  machine translation}.
\newblock \bibinfo{journal}{\emph{arXiv preprint arXiv:1508.04025}}
  (\bibinfo{year}{2015}).
\newblock


\bibitem[\protect\citeauthoryear{Mikolov, Sutskever, Chen, Corrado, and
  Dean}{Mikolov et~al\mbox{.}}{2013}]%
        {mikolov2013distributed}
\bibfield{author}{\bibinfo{person}{Tomas Mikolov}, \bibinfo{person}{Ilya
  Sutskever}, \bibinfo{person}{Kai Chen}, \bibinfo{person}{Greg~S Corrado},
  {and} \bibinfo{person}{Jeff Dean}.} \bibinfo{year}{2013}\natexlab{}.
\newblock \showarticletitle{Distributed representations of words and phrases
  and their compositionality}. In \bibinfo{booktitle}{\emph{Advances in neural
  information processing systems}}. \bibinfo{pages}{3111--3119}.
\newblock


\bibitem[\protect\citeauthoryear{Pandey and Jha}{Pandey and Jha}{2016}]%
        {pandey2016error}
\bibfield{author}{\bibinfo{person}{Rajneesh~Kumar Pandey} {and}
  \bibinfo{person}{Girish~Nath Jha}.} \bibinfo{year}{2016}\natexlab{}.
\newblock \showarticletitle{Error analysis of sahit-a statistical
  Sanskrit-Hindi translator}.
\newblock \bibinfo{journal}{\emph{Procedia Computer Science}}
  \bibinfo{volume}{96} (\bibinfo{year}{2016}), \bibinfo{pages}{495--501}.
\newblock


\bibitem[\protect\citeauthoryear{Reddi, Kale, and Kumar}{Reddi
  et~al\mbox{.}}{2019}]%
        {reddi2019convergence}
\bibfield{author}{\bibinfo{person}{Sashank~J Reddi}, \bibinfo{person}{Satyen
  Kale}, {and} \bibinfo{person}{Sanjiv Kumar}.}
  \bibinfo{year}{2019}\natexlab{}.
\newblock \showarticletitle{On the convergence of adam and beyond}.
\newblock \bibinfo{journal}{\emph{arXiv preprint arXiv:1904.09237}}
  (\bibinfo{year}{2019}).
\newblock


\bibitem[\protect\citeauthoryear{Schuster, Johnson, and Thorat}{Schuster
  et~al\mbox{.}}{2016}]%
        {Schuster2016zero}
\bibfield{author}{\bibinfo{person}{Mike Schuster}, \bibinfo{person}{Melvin
  Johnson}, {and} \bibinfo{person}{Nikhil Thorat}.}
  \bibinfo{year}{2016}\natexlab{}.
\newblock \showarticletitle{Zero-shot translation with Google\textquotesingle s
  multilingual neural machine translation system}.
\newblock \bibinfo{journal}{\emph{Google Research Blog}}  \bibinfo{volume}{22}
  (\bibinfo{year}{2016}).
\newblock


\bibitem[\protect\citeauthoryear{Schuster and Paliwal}{Schuster and
  Paliwal}{1997}]%
        {schuster1997bidirectional}
\bibfield{author}{\bibinfo{person}{Mike Schuster} {and}
  \bibinfo{person}{Kuldip~K Paliwal}.} \bibinfo{year}{1997}\natexlab{}.
\newblock \showarticletitle{Bidirectional recurrent neural networks}.
\newblock \bibinfo{journal}{\emph{IEEE Transactions on Signal Processing}}
  \bibinfo{volume}{45}, \bibinfo{number}{11} (\bibinfo{year}{1997}),
  \bibinfo{pages}{2673--2681}.
\newblock


\bibitem[\protect\citeauthoryear{Shterionov, Nagle, Casanellas, Superbo, and
  O\textquotesingle~Dowd}{Shterionov et~al\mbox{.}}{2017}]%
        {shterionov2017empirical}
\bibfield{author}{\bibinfo{person}{Dimitar Shterionov}, \bibinfo{person}{Pat
  Nagle}, \bibinfo{person}{Laura Casanellas}, \bibinfo{person}{Riccardo
  Superbo}, {and} \bibinfo{person}{Tony O\textquotesingle~Dowd}.}
  \bibinfo{year}{2017}\natexlab{}.
\newblock \showarticletitle{Empirical evaluation of NMT and PBSMT quality for
  large-scale translation production}. In \bibinfo{booktitle}{\emph{Proceedings
  of the Annual Conference of the European Association for Machine Translation
  {(EAMT)}: User Track}}.
\newblock


\bibitem[\protect\citeauthoryear{Singh, Kumar, and Chana}{Singh
  et~al\mbox{.}}{2019}]%
        {singh2019improving}
\bibfield{author}{\bibinfo{person}{Muskaan Singh}, \bibinfo{person}{Ravinder
  Kumar}, {and} \bibinfo{person}{Inderveer Chana}.}
  \bibinfo{year}{2019}\natexlab{}.
\newblock \showarticletitle{Improving Neural Machine Translation Using
  Rule-Based Machine Translation}. In \bibinfo{booktitle}{\emph{2019 7th
  International Conference on Smart Computing \& Communications {(ICSCC)}}}.
  IEEE, \bibinfo{pages}{1--5}.
\newblock


\bibitem[\protect\citeauthoryear{Sutskever, Vinyals, and Le}{Sutskever
  et~al\mbox{.}}{2014}]%
        {sutskever2014sequence}
\bibfield{author}{\bibinfo{person}{Ilya Sutskever}, \bibinfo{person}{Oriol
  Vinyals}, {and} \bibinfo{person}{Quoc~V Le}.}
  \bibinfo{year}{2014}\natexlab{}.
\newblock \showarticletitle{Sequence to sequence learning with neural
  networks}. In \bibinfo{booktitle}{\emph{Advances in neural information
  processing systems}}. \bibinfo{pages}{3104--3112}.
\newblock


\bibitem[\protect\citeauthoryear{Valmiki.iitk.ac.in}{Valmiki.iitk.ac.in}{2019}]%
        {sa-en2}
\bibfield{author}{\bibinfo{person}{Valmiki.iitk.ac.in}.}
  \bibinfo{year}{2019}\natexlab{}.
\newblock \bibinfo{booktitle}{\emph{Content | Valmiki Ramayanam}}.
\newblock
\urldef\tempurl%
\url{https://www.valmiki.iitk.ac.in/content?language=dv&field_kanda_tid=1&field_sarga_value=1&field_sloka_value=1}
\showURL{%
Retrieved November 23, 2019 from \tempurl}


\end{thebibliography}
\nocite{singh2019improving}
%
%
%
%
%
%
%
%
\vspace{-8ex}
\end{document}